\documentclass[final,5p,times,twocolumn,nopreprintline]{elsarticle}

\usepackage{amssymb}
\usepackage{amsmath}

\usepackage{graphicx}
\usepackage{subfigure}
\usepackage{booktabs}
\usepackage{multirow}

\begin{document}

\begin{frontmatter}



\title{Data Augmentation in Time Series Forecasting through Inverted Framework}


\author[inst1,inst2]{Hongming Tan}
\ead{thm22@mails.tsinghua.edu.cn}

\author[inst1,inst2]{Ting Chen}
\ead{chent22@mails.tsinghua.edu.cn}

\author[inst1,inst2]{Ruochong Jin}
\ead{jrc23@mails.tsinghua.edu.cn}

\author[inst1,inst2]{Wai Kin (Victor) Chan\corref{cor1}}
\ead{chanw@sz.tsinghua.edu.cn}

\cortext[cor1]{Corresponding author.}

\affiliation[inst1]{organization={Shenzhen International Graduate School, Tsinghua University},
            city={Shenzhen},
            country={China}}
\affiliation[inst2]{organization={Pengcheng Laboratory},
            city={Shenzhen},
            country={China}}

\begin{abstract}
Currently, iTransformer is one of the most popular and effective models for multivariate time series (MTS) forecasting. Thanks to its inverted framework, iTransformer effectively captures multivariate correlation. However, the inverted framework still has some limitations. It diminishes temporal interdependency information, and introduces noise in cases of nonsignificant variable correlation. To address these limitations, we introduce a novel data augmentation method on inverted framework, called DAIF. Unlike previous data augmentation methods, DAIF stands out as the first real-time augmentation specifically designed for the inverted framework in MTS forecasting. We first define the structure of the inverted sequence-to-sequence framework, then propose two different DAIF strategies, Frequency Filtering and Cross-variation Patching to address the existing challenges of the inverted framework. Experiments across multiple datasets and inverted models have demonstrated the effectiveness of our DAIF.
Our codes are available at https://github.com/Travistan123/time-series-daif.
\end{abstract}



    
    


\end{frontmatter}



\section{Introduction}
\label{sec:intro}
Multivariate time series (MTS) forecasting is a pivotal topic in time series analysis with wide-ranging applications.
Recently, deep learning sequence-to-sequence (Seq2Seq) MTS forecasting models range from MLP-based \cite{zeng2023transformers,wangtimemixer} 
to RNN-based \cite{cho2014learning,lai2018modeling,salinas2020deepar} and 
Transformer-based models \cite{vaswani2017attention,li2019enhancing,wu2021autoformer,zhang2023crossformer,liu2024itransformer}. 
The structured state space sequence model (SSM) Mamba \cite{gu2023mamba} has also attracted research interest.
Importantly, Transformer-based models for time series can be categorized into three levels: point-wise, patch-wise, and series-wise representations \cite{wang2024deep}. Vanilla Transformer applies a point-wise embedding that often overlooks local semantics. PatchTST \cite{nie2023time} introduces patch-wise tokens to capture dependencies across segments. 
Further, the series-wise model iTransformer \cite{liu2024itransformer} embeds the entire series for each variate as a token. 
Unlike the vanilla Transformer, iTransformer adopts an inverted structure that captures multivariate correlations more effectively and leads to improved forecasting accuracy.
We generalize this architecture as the \textbf{inverted Seq2Seq framework} in time series analysis.

\begin{figure}
  \centering
  \includegraphics[width=1\linewidth]{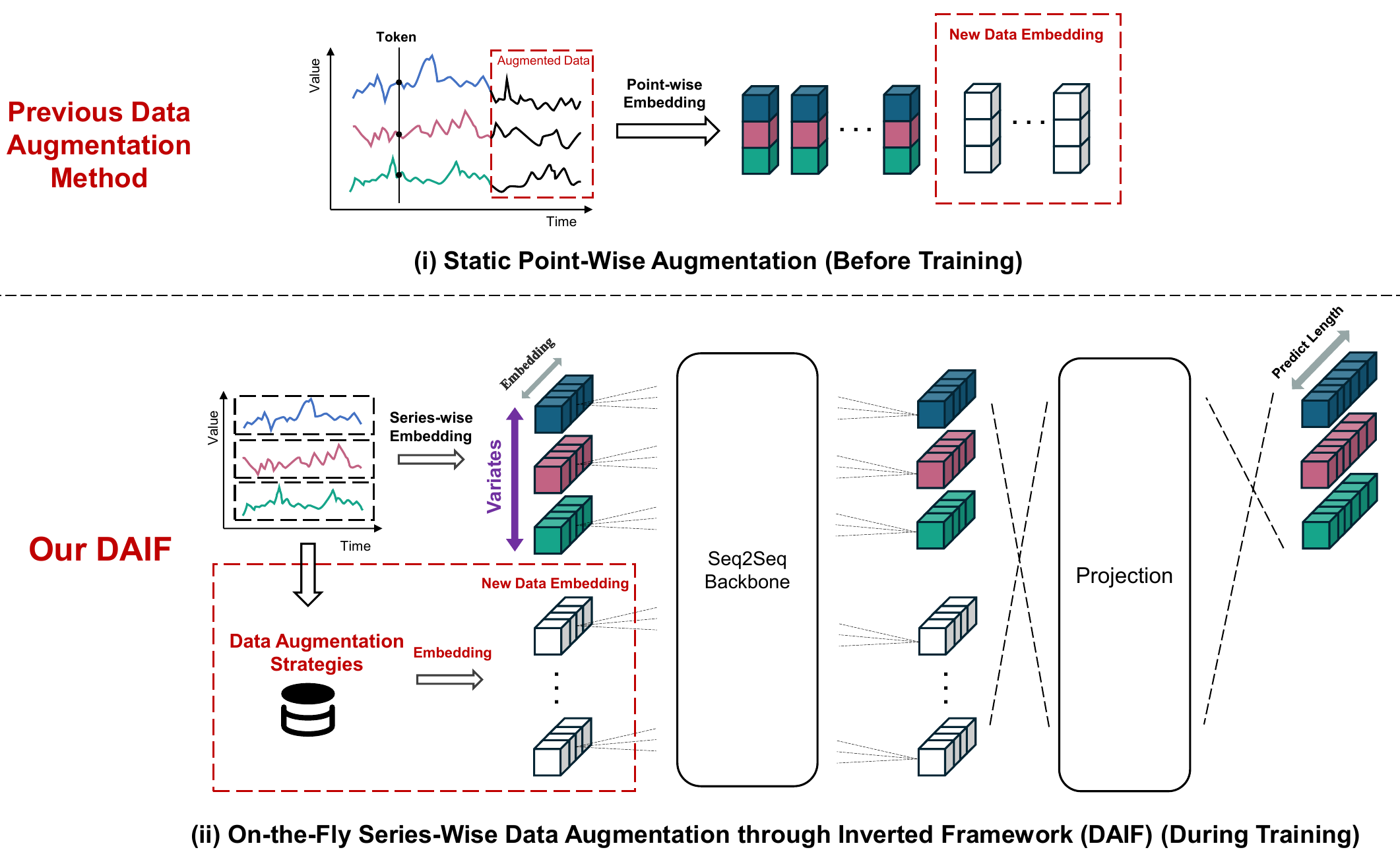}
  \caption{Comparison between (i) previous point-wise augmentation method and (ii) our on-the-fly DAIF framework. As shown in the example, DAIF processes a 3-variate time series by applying series-wise embedding to each original variable and introducing multiple on-the-fly augmented embeddings during training.
  }
  \label{data_aug_overview}
\end{figure}

The inverted Seq2Seq framework effectively captures multivariate correlations for variate-centric representations but encounters two primary challenges. 
First, this framework overlooks local temporal correlations within the series, as it relies only on series-wise representations without the patch-wise perspective. Second, the inverted framework may introduce noise when focusing on multivariate correlations, but there is no significant correlation between variates \cite{wang2024timexer}.
We observe that the form of the inverted framework enables us to propose a new perspective on data augmentation. As shown in Fig. \ref{data_aug_overview}(ii), during training and testing, embedding each variate independently as a token ensures that adding new tokens does not disrupt the integrity of existing sequences. Instead, tokens generated from the original series can augment the existing sequence representations.

In this paper, we propose on-the-fly \textbf{D}ata \textbf{A}ugmentation through \textbf{I}nverted \textbf{F}ramework (DAIF) to address existing deficiencies in the inverted Seq2Seq framework. 
Previous time series data augmentation methods \cite{wen2021time, gao2020robusttad,bandara2021improving,hu2020datsing} typically use point-wise augmentation to increase labeled training data and set up transfer learning systems, as shown in Fig. \ref{data_aug_overview}(i). However, there is a lack of research on lightweight, real-time data augmentation methods that organically integrate with existing model frameworks.
Fig. \ref{data_aug_overview}(ii) shows that the inverted Seq2Seq framework consists of series-wise embedding, a Seq2Seq backbone (such as Transformer or Mamba blocks), and a linear projection for aligning the output prediction length.
DAIF applies multivariate time series augmentation strategies to extract valuable information from the original series and generate new data embeddings. These new embeddings are inputs to the inverted models together with the original series-wise embeddings. The number of tokens remains constant throughout the Seq2Seq backbone, and the final output retains only the number of tokens present in the original series input.

Specifically, in DAIF, we explore two data augmentation strategies.
(i) To address weakened learning of internal correlations within series, we introduce \textbf{Cross-variation Patching}. This time series representation approach constructs cross-variation temporal tokens like point-wise approaches but patches multiple timestamps simultaneously.
(ii) To tackle the issue of insignificant correlations between variates, we propose \textbf{Frequency Filtering}. This method uses Fast Fourier Transform (FFT) and inverse FFT to construct denoised series that preserve the top-K amplitude frequencies \cite{wu2023timesnet}. This preservation ensures alignment with the key features of the raw series.

The main contributions of this paper are as follows:
\begin{itemize}
    \item To the best of our knowledge, DAIF is the first on-the-fly data augmentation method for the inverted framework in time series forecasting.
    \item We conduct experiments to demonstrate the effectiveness of two DAIF strategies, Frequency Filtering and Cross-variation Patching.
    \item We generalize the inverted framework, originally limited to Transformers, to a Seq2Seq structure applicable to MLP and RNN models.
\end{itemize}

\section{Related Work}
In this section, we review three categories of related work pertinent to our DAIF: non-transformer-based methods, transformer-based methods, and data augmentation methods.

\subsection{Non-Transformer-based Methods}
Multivariate time series (MTS) forecasting has seen substantial progress across various methodologies. Early works mainly adopted classical statistical methods such as ARIMA and GARCH \cite{box2015time}. 
Recent advancements shifted the paradigm towards deep learning techniques. 
Notably, multilayer perceptron (MLP) based models such as DLinear \cite{zeng2023transformers}, RLinear \cite{li2023revisiting} and TimeMixer \cite{wangtimemixer} achieve competitive results with simple architectures.
Besides, recurrent neural network (RNN) based models, ranging from LSTM \cite{hochreiter1997long, zhao2017lstm} to more variants such as LSTNet \cite{lai2018modeling} and DeepAR \cite{salinas2020deepar}, have gained prominence for their ability to capture temporal dependencies.
Representative CNN-based models include TCN \cite{franceschi2019unsupervised} and TimesNet \cite{wu2023timesnet}.
Recently, structured state space models (SSMs), such as Mamba \cite{gu2023mamba,ahamed2024timemachine}, have emerged as a RNN-like approach for efficient sequence modeling.

\subsection{Transformer-based Methods}
Transformer-based methods enhance MTS forecasting by effectively modeling long-range dependencies through attention mechanisms \cite{vaswani2017attention}.
These models can be classified into three categories based on their embedding representations: point-wise, patch-wise, and series-wise \cite{wang2024deep}. 
Vanilla Transformer and several early Transformer-based models, such as LogSparse \cite{li2019enhancing} and Informer \cite{zhou2021informer}, use point-wise embeddings that represent each time step individually. Point-wise representations often fails to capture local semantic patterns within temporal neighborhoods.
Patch-wise models, such as PatchTST \cite{nie2023time}, Autoformer \cite{wu2021autoformer} and Crossformer \cite{zhang2023crossformer}, improve upon this by segmenting data into patches to capture inter-patch dependencies. The recently proposed series-wise model iTransformer \cite{liu2024itransformer} takes this further by embedding each variate's entire series as a token, effectively capturing multivariate correlations. 
This study defines the series-wise representation approach as an inverted Seq2Seq framework and introduces DAIF, a data augmentation framework specifically designed to improve its effectiveness.

\subsection{Data Augmentation Methods}

Data augmentation methods for time series can be broadly categorized into three groups: time domain, frequency domain, and time-frequency domain approaches. Time domain methods include operations such as jittering, scaling, and rotating \cite{rashid2019times, park2019specaugment}. Frequency domain techniques involve transformations like phase perturbations (APP) \cite{gao2020robusttad}, the amplitude-adjusted Fourier transform (AAFT), and the iterated AAFT (IAAFT) \cite{schreiber2000surrogate}. Time-frequency domain methods operate across both temporal and spectral dimensions, such as the Short-Time Fourier Transform (STFT) \cite{steven2018feature} and SpecAugment \cite{park2019specaugment}. 
Most existing approaches apply point-wise augmentation before training to enlarge labeled datasets.
In contrast, our proposed DAIF framework performs on-the-fly augmentation in the time-frequency domain, and is the first such method specifically designed for the series-wise inverted framework.

\section{Method}
\label{sec:method}
In this section, we first introduce DAIF as a generic data augmentation inverted framework. Subsequently, we propose two specific data augmentation strategies within DAIF.

\subsection{DAIF}
\label{ssec:DAIF}
Given a MTS \( \mathbf{X} \in \mathbb{R}^{T \times N} \) with \( T \) time steps and \( N \) variates, the series is represented as \( \mathbf{X} = [\mathbf{x}_1, \mathbf{x}_2, \ldots, \mathbf{x}_T]\), with \( \mathbf{x}_t \in \mathbb{R}^N \). 
The forecasting objective is to predict \( S \) future time steps: \( \hat{\mathbf{Y}} = [\mathbf{y}_1, \mathbf{y}_2, \ldots, \mathbf{y}_S] \in \mathbb{R}^{S \times N} \).
Data Augmentation through Inverted Framework (DAIF) applies in real time to \( \mathbf{X} \) to produce augmented data \( \overline{\mathbf{X}} \in \mathbb{R}^{J \times M} \). The quantity \( M \) and length \( J \) of \( \overline{\mathbf{X}} \) vary according to different specific DAIF strategies. DAIF is formulated as follows:

\begin{align}
\overline{\mathbf{X}} &= \text{Augmentation}(\mathbf{X}), \\
\mathbf{h}_{n}^0 &= \text{Embedding}(\mathbf{X}_{:,n}), \quad n = 1, \ldots, N, \label{eq:embedding1} \\
\overline{\mathbf{h}}_{m}^0 &= \text{Embedding}(\overline{\mathbf{X}}_{:,m}), \quad m = 1, \ldots, M, \label{eq:embedding2} \\
\mathbf{H}^{l+1} &= \text{Seq2SeqBlock}(\mathbf{H}^l), \quad l = 0, \ldots, L-1, \\
\hat{\mathbf{Y}} &= \text{Projection}(\mathbf{H}^{L})_{1:N},
\end{align}

where each \( \mathbf{X}_{:,n} \) of \( \mathbf{X} \) represent the whole series for the \( n \)-th variate, and each \( \overline{\mathbf{X}}_{:,m} \) of \(\overline{\mathbf{X}}\) represent the \( m \)-th augmented token.
In Eq.(\ref{eq:embedding1}) and Eq.(\ref{eq:embedding2}), \(\text{Embedding}(\mathbf{X}_{:,n})\) maps from \( \mathbb{R}^{T} \) to \( \mathbb{R}^D \), and \(\text{Embedding}(\overline{\mathbf{X}})\) maps from \( \mathbb{R}^{J} \) to \( \mathbb{R}^D \).
\( \mathbf{H} = \{\mathbf{h}_1, ..., \mathbf{h}_{N}, \overline{\mathbf{h}}_{1}, ..., \overline{\mathbf{h}}_{M}\} \in \mathbb{R}^{(N+M) \times D} \) consists of \( N + M \) embeddings with \( D \) dimension. Superscript \(l\) denotes the Seq2Seq layer index.
\( \text{Seq2SeqBlock}( \cdot ) \) maintains the total count of \( N + M \) tokens between input and output.
Finally, \(\text{Projection}(\mathbf{H}^{L})_{1:N}\) maps from \( \mathbb{R}^{(N+M) \times D} \) to \( \mathbb{R}^{(N+M) \times S} \) and then selects only the first \(N\) tokens as output with \( \mathbb{R}^{N \times S} \).

In DAIF, multiple Seq2Seq blocks process both original and augmented data embeddings, each in \( \mathbb{R}^{D} \) dimensions, to learn multivariate correlations of all data and also to understand the representations of augmented data.

\subsection{Cross-variation Patching}
\label{ssec:patching}

\begin{figure}[t]
  \centering
  \includegraphics[width=1\linewidth]{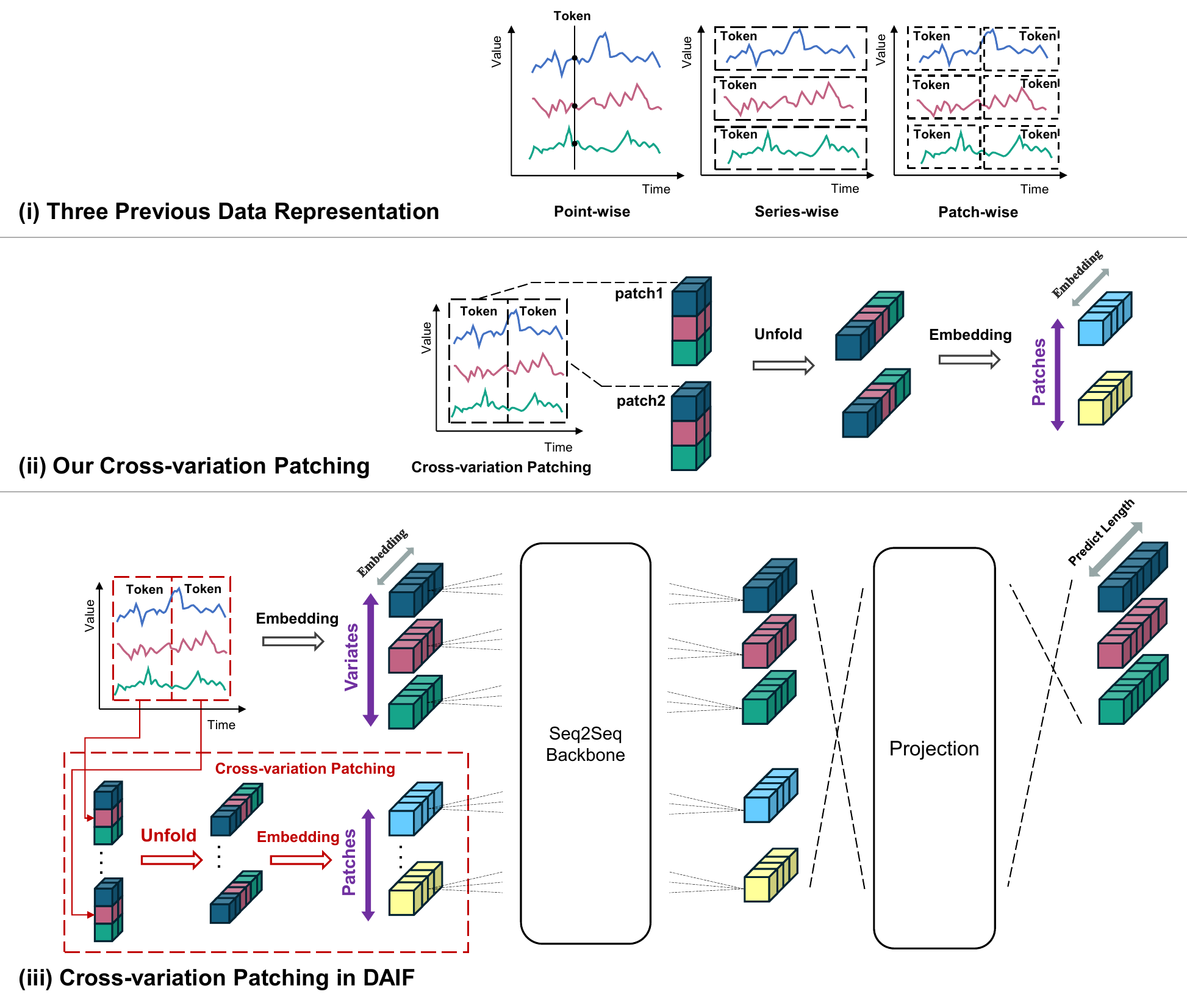}
  \caption{Overview of (i) three previous data representation, (ii) Cross-variation Patching implementation, and (iii) an example of Cross-variation Patching in DAIF.}
  \label{patch_overview}
\end{figure}

As shown in Fig. \ref{patch_overview}, unlike existing point-wise, patch-wise, and series-wise data representation, we propose a novel time series data representation called Cross-variation Patching. From the perspective of Cross-variation Patching, we treat multivariate series as univariate series and patch it into multiple tokens. 
Specifically, Cross-variation Patching is calculated though reshaping \(\mathbf{X}\):
\begin{align}
\{Patch_k\} &= \text{CrossVariationPatch}(\mathbf{X}), \quad k = 1, \ldots, M, \\
\overline{\mathbf{X}} &= \text{Unfolding}(\{Patch_k\}). \label{eq:unfolding}
\end{align}
Given the patch length \( P \), each \(Patch_k = \mathbf{X}_{(k-1)P:kP, :} \in \mathbb{R}^{P \times N}\) represents the \(k\)-th cross-variation patch extracted from \(\mathbf{X}\). \( M \) is the total number of patches, calculated as \( M = \lfloor \frac{T}{P} \rfloor \).
Then in Eq.(\ref{eq:unfolding}), the unfolding operation flattens each \(\mathbf{X}_{(k-1)P:kP, :}\) into a one-dimensional token of length \(J\) for subsequent embedding.
Finally, we get augmented \(\overline{\mathbf{X}}\) with \(M\) total tokens and each token length \(J = PN\).

The benefits of this representation are: (i) it complements the series-wise representation by providing cross-variation local semantic information in temporal view, and (ii) it applies a lightweight data augmentation, generating only a small number of patches (\(M\)) from a MTS.

\subsection{Frequency Filtering}
\label{ssec:frequency}
\begin{figure}[t]
  \centering
  \includegraphics[width=1\linewidth]{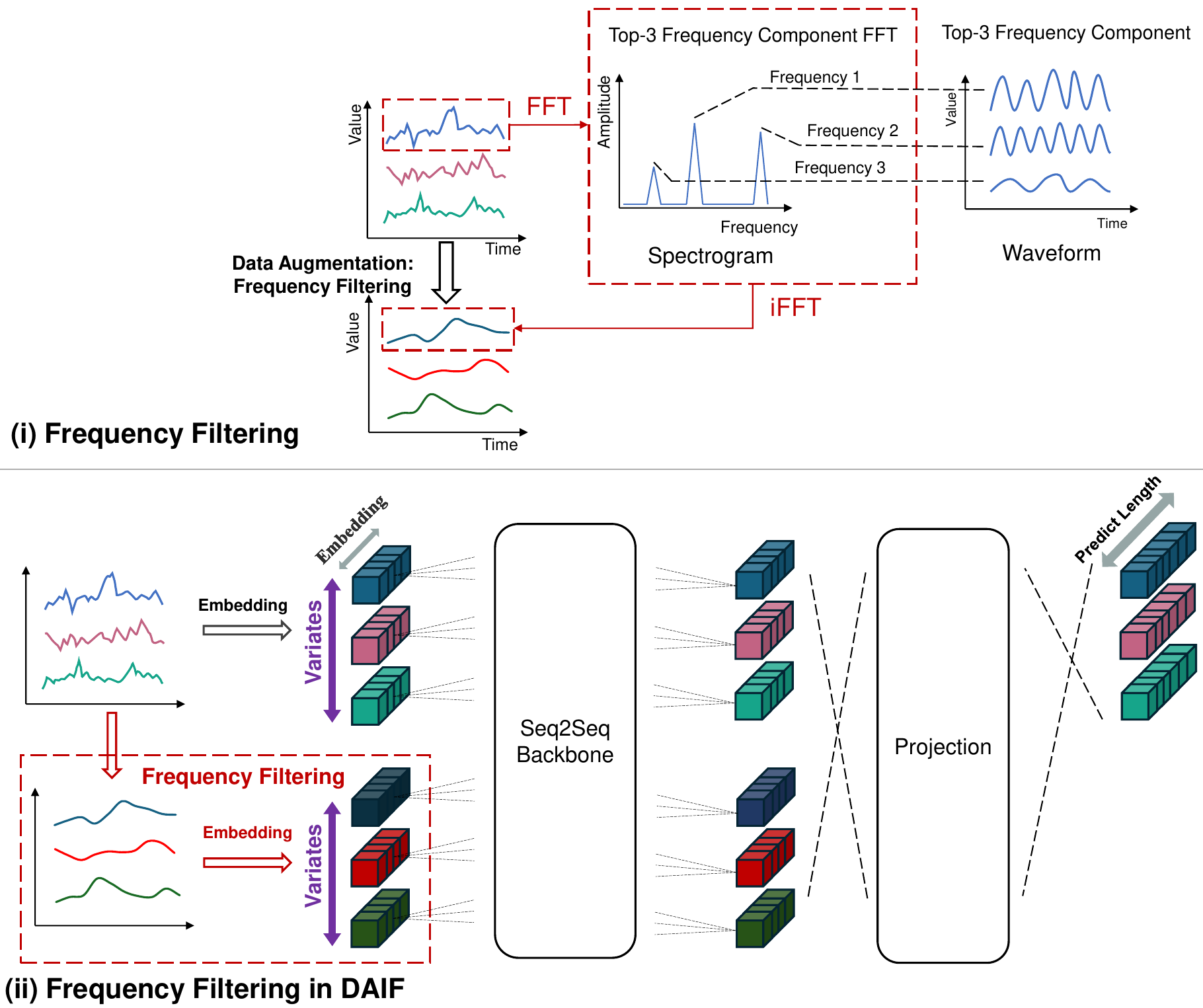}
  \caption{Example of (i) top-3 Frequency Filtering implementation, and (ii) Frequency Filtering in DAIF.}
  \label{fft_overview}
\end{figure}

The second DAIF strategy we introduce is the Frequency Filtering method. Fig. \ref{fft_overview} (i) shows that this approach initially utilizes Fast Fourier Transform (FFT) to convert the original time-series data into the frequency domain and preserves the top-K amplitude frequency components like TimesNet \cite{wu2023timesnet}. FFT decomposes the time series into frequencies to identify crucial spectral features.
Subsequently, the inverse Fast Fourier Transform (iFFT) converts these frequency-domain components back into the time domain to reconstruct the series with an emphasis on high-amplitude frequencies.
We summarize the Frequency Filtering process as follows:
\begin{align}
\{f_1, \ldots, f_k\} &= \text{argTopK}(\text{Amp}(\text{FFT}(\mathbf{X}_{:,n}))), \\
\overline{\mathbf{X}}_{:,n} &= \text{iFFT}(\{f_1, \ldots, f_k\}),
\end{align}
where \( \text{FFT}( \cdot ) \) transforms each univariate series \(\mathbf{X}_{:,n}\) from time to frequency domain, \( \text{Amp}( \cdot ) \) calculates the amplitude of each frequency component, and \( \text{argTopK}( \cdot ) \) preserves \(k\) frequency components with top-K highest amplitudes.
\( \text{iFFT}( \cdot ) \) converts these \(k\) frequency components into a new time series \( \overline{\mathbf{X}}_{:,n} \), which highly correlates with \( \mathbf{X}_{:,n} \).
The denoised multivariate series \(\overline{\mathbf{X}} \in \mathbb{R}^{T \times N}\) always has the same dimensions of \(\mathbf{X}\).

Frequency Filtering further improves the multivariate correlation capabilities of the inverted framework by constructing new time series that incorporate the significant frequency components of the original series.
From Fig. \ref{fft_overview} we can see that the new series keep the main trends of the original series.

\section{Experiment}

\subsection{Experiment Setup}
\textbf{Datasets.} 
We conducted experiments on six long-term forecasting benchmarks: Electricity, Weather, Traffic, Exchange \cite{wu2021autoformer}, Solar Energy \cite{lai2018modeling}, and ETT datasets (4 subsets) \cite{zhou2021informer}, and on the short-term forecasting dataset PEMS (4 subsets) \cite{liu2022scinet}.
Our experiments comprehensively cover representative time series forecasting datasets from various domains.

\textbf{Baselines and Setup.} 
We selected three inverted models as baselines to evaluate the performance improvement of our DAIF: (i) iTransformer, (ii) inverted Mamba, and (iii) inverted multilayer perceptron (MLP). 
iTransformer \cite{liu2024itransformer} represents the state-of-the-art Transformer-based model. We also implemented the inverted versions of the advanced RNN-based Mamba \cite{gu2023mamba} model and the classical MLP.
 
All models were developed using the Time-Series-Library \cite{wang2024deep} and maintained consistent parameter settings across comparative experiments.
Specifically, single-layer Mamba and two-layer MLP with GELU activation function \cite{wangtimemixer} were used in the experiments. 
For datasets, we employed prediction length \(S \in \{12, 24\}\) for PEMS and \(S \in \{96, 192, 336, 720\}\) for the other datasets.
The lookback length was fixed at \(T = 96\). 
We used Mean Square Error (MSE) and Mean Absolute Error (MAE) as evaluation metrics.
All experimental results were averaged over three runs.

\begin{table}[ht]
\centering
\caption{Performance improvement of the inverted framework.}
\label{tab:inverted_enhancement}
\setlength{\tabcolsep}{2pt}
\begin{tabular}{@{}c@{}|c|c c|c c|c c@{}}
\toprule
\multicolumn{2}{c|}{Models}               & \multicolumn{2}{c|}{Transformer} & \multicolumn{2}{c|}{Mamba}       & \multicolumn{2}{c}{MLP}         \\ \cmidrule(lr){3-4}\cmidrule(lr){5-6}\cmidrule(l){7-8}
\multicolumn{2}{c|}{Metric}               & MSE            & MAE            & MSE            & MAE            & MSE            & MAE            \\ \midrule
\multirow{2}{*}{Electricity} & Original  & 0.277          & 0.372          & 0.235          & 0.334          & 0.208          & 0.312          \\ 
                             & \textbf{+Inverted} & \textbf{0.178} & \textbf{0.270} & \textbf{0.222} & \textbf{0.307} & \textbf{0.204} & \textbf{0.285} \\ \midrule
\multirow{2}{*}{Traffic}     & Original  & 0.665          & 0.363          & 0.789          & 0.454          & 0.675          & 0.396          \\ 
                             & \textbf{+Inverted} & \textbf{0.428} & \textbf{0.282} & \textbf{0.574} & \textbf{0.385} & \textbf{0.500} & \textbf{0.319} \\ \midrule
\multirow{2}{*}{Weather}     & Original  & 0.657          & 0.572          & 0.279          & 0.301          & 0.274          & 0.299          \\ 
                             & \textbf{+Inverted} & \textbf{0.258} & \textbf{0.279} & \textbf{0.264} & \textbf{0.286} & \textbf{0.273} & \textbf{0.290} \\ \bottomrule
\end{tabular}
\end{table}

\subsection{Effectiveness of Inverted Framwork}
Table \ref{tab:inverted_enhancement} shows that all three inverted models outperform their original models on three benchmarks. 
It indicates that the inverted framework is applicable not only to Transformer-based models but also can be extended to other backbone models.
Additionally, the number of variates in each dataset may also influence the forecasting results. 
For instance, the average MSE performance improvement of the inverted framework on the 862-variate Traffic dataset is 29.6\%, which is more pronounced compared to the 321-variate Electricity dataset and the 21-variate Weather dataset.
These results strongly demonstrate the effectiveness of the inverted Seq2Seq framework that forms the basis of DAIF.

\subsection{Main Results of DAIF}


\begin{table*}[!ht]
\centering
\caption{Performance comparison of inverse models with and without DAIF. \textbf{Bold} highlights the best result and \underline{underline} highlights the second best result for each model. Each Results within each dataset are averaged across all prediction lengths.
}
\label{tab:DAIF_all_results}
\resizebox{\textwidth}{!}{%
\begin{tabular}{@{}c|cccccc|cccccc|cccccc@{}}
\toprule
\multirow{2}{*}{Models} & \multicolumn{6}{c|}{iTransformer} & \multicolumn{6}{c|}{inverted Mamba} & \multicolumn{6}{c}{inverted MLP} \\ 
\cmidrule(lr){2-7} \cmidrule(lr){8-13} \cmidrule(lr){14-19}
& \multicolumn{2}{c}{w/o aug} & \multicolumn{2}{c}{\textbf{w/ DAIF-CvP}} & \multicolumn{2}{c|}{\textbf{w/ DAIF-FF}} & \multicolumn{2}{c}{w/o aug} & \multicolumn{2}{c}{\textbf{w/ DAIF-CvP}} & \multicolumn{2}{c|}{\textbf{w/ DAIF-FF}} & \multicolumn{2}{c}{w/o aug} & \multicolumn{2}{c}{\textbf{w/ DAIF-CvP}} & \multicolumn{2}{c}{\textbf{w/ DAIF-FF}} \\ 
\cmidrule(lr){2-3} \cmidrule(lr){4-5} \cmidrule(lr){6-7} \cmidrule(lr){8-9} \cmidrule(lr){10-11} \cmidrule(lr){12-13} \cmidrule(lr){14-15} \cmidrule(lr){16-17} \cmidrule(lr){18-19}
Metric & MSE & MAE & MSE & MAE & MSE & MAE & MSE & MAE & MSE & MAE & MSE & MAE & MSE & MAE & MSE & MAE & MSE & MAE \\
\midrule
Electricity & 0.178 & 0.270 & \underline{0.168} & \underline{0.263} & \textbf{0.165} & \textbf{0.261} & 0.222 & 0.307 & \textbf{0.220} & \textbf{0.306} & \underline{0.222} & 0.309 & 0.204 & 0.285 & \textbf{0.203} & \textbf{0.284} & \underline{0.204} & \underline{0.284} \\
Weather & 0.258 & 0.279 & \underline{0.247} & \textbf{0.274} & \textbf{0.246} & \underline{0.274} & 0.264 & 0.286 & \textbf{0.250} & \textbf{0.277} & \underline{0.263} & \underline{0.285} & 0.273 & 0.290 & \textbf{0.271} & \textbf{0.289} & \underline{0.273} & \underline{0.290} \\
ETTh1 & 0.454 & 0.447 & \textbf{0.444} & \textbf{0.439} & \underline{0.446} & \underline{0.441} & 0.468 & 0.452 & \underline{0.467} & \underline{0.451} & \textbf{0.465} & \textbf{0.450} & 0.454 & 0.439 & \textbf{0.453} & \textbf{0.439} & \underline{0.454} & \underline{0.439} \\
ETTh2 & 0.383 & 0.407 & \underline{0.381} & \underline{0.406} & \textbf{0.379} & \textbf{0.404} & \underline{0.389} & \underline{0.406} & \textbf{0.387} & \textbf{0.405} & 0.390 & 0.407 & 0.376 & \underline{0.401} & \underline{0.376} & 0.402 & \textbf{0.375} & \textbf{0.401} \\
ETTm1 & 0.407 & 0.410 & \underline{0.399} & \underline{0.406} & \textbf{0.396} & \textbf{0.404} & 0.390 & 0.401 & \textbf{0.388} & \textbf{0.399} & \underline{0.389} & \underline{0.400} & 0.407 & 0.407 & \underline{0.407} & \underline{0.407} & \textbf{0.406} & \textbf{0.406} \\
ETTm2 & 0.288 & 0.332 & \textbf{0.282} & \textbf{0.327} & \underline{0.284} & \underline{0.328} & 0.282 & 0.326 & \textbf{0.279} & \textbf{0.324} & \underline{0.282} & \underline{0.326} & 0.285 & 0.329 & \underline{0.281} & \underline{0.326} & \textbf{0.280} & \textbf{0.326} \\
Traffic & 0.428 & 0.282 & \textbf{0.417} & \underline{0.278} & \underline{0.420} & \textbf{0.276} & 0.574 & 0.385 & \underline{0.573} & \underline{0.385} & \textbf{0.573} & \textbf{0.385} & 0.501 & 0.319 & \textbf{0.494} & \textbf{0.316} & \underline{0.500} & \underline{0.318} \\
Solar-Energy & \underline{0.233} & \underline{0.261} & \textbf{0.232} & \textbf{0.260} & 0.235 & 0.263 & 0.305 & 0.316 & \textbf{0.291} & \textbf{0.318} & \underline{0.303} & \underline{0.319} & 0.275 & 0.293 & \underline{0.271} & \underline{0.290} & \textbf{0.269} & \textbf{0.289} \\
Exchange & \underline{0.360} & \textbf{0.403} & \textbf{0.359} & \underline{0.404} & 0.360 & 0.405 & 0.370 & 0.409 & \textbf{0.370} & \textbf{0.409} & \underline{0.370} & \underline{0.409} & 0.350 & 0.397 & \textbf{0.349} & \textbf{0.397} & \underline{0.350} & \underline{0.397} \\
PEMS03 & 0.082 & 0.188 & \textbf{0.072} & \textbf{0.178} & \underline{0.073} & \underline{0.179} & 0.113 & 0.224 & \underline{0.102} & \underline{0.215} & \textbf{0.101} & \textbf{0.214} & 0.123 & 0.233 & \textbf{0.122} & \textbf{0.231} & \underline{0.122} & \underline{0.232} \\
PEMS04 & 0.087 & 0.194 & \textbf{0.079} & \textbf{0.187} & \underline{0.079} & \underline{0.188} & 0.133 & 0.244 & \underline{0.123} & \underline{0.235} & \textbf{0.122} & \textbf{0.235} & 0.120 & 0.233 & \textbf{0.107} & \textbf{0.217} & \underline{0.107} & \underline{0.218} \\
PEMS07 & 0.078 & 0.178 & \textbf{0.067} & \textbf{0.167} & \underline{0.069} & \underline{0.169} & 0.111 & 0.223 & \textbf{0.107} & \textbf{0.220} & \underline{0.109} & \underline{0.221} & 0.104 & 0.215 & \textbf{0.094} & \textbf{0.204} & \underline{0.095} & \underline{0.205} \\
PEMS08 & 0.097 & 0.201 & \underline{0.092} & \underline{0.197} & \textbf{0.090} & \textbf{0.195} & 0.130 & 0.237 & \textbf{0.118} & \textbf{0.227} & \underline{0.119} & \underline{0.227} & 0.147 & 0.252 & \underline{0.147} & \underline{0.252} & \textbf{0.146} & \textbf{0.251} \\
\bottomrule
\end{tabular}%
}
\end{table*}

\begin{table*}[ht]
\centering
\caption{Performance comparison of DAIF and other competitive models. \textbf{Bold} highlights the best result and \underline{underline} highlights the second best result for each model.}
\label{tab:DAIF_more_comparisons}
\resizebox{\textwidth}{!}{
\begin{tabular}{@{}l
|cc 
|cc 
|cc 
|cc 
|cc 
|cc 
|cc 
|cc 
|cc 
|cc 
@{}}
\toprule
\multirow{2}{*}{Dataset} 
  & \multicolumn{2}{c|}{\textbf{DAIF-CvP}} 
  & \multicolumn{2}{c|}{\textbf{DAIF-FF}}
  & \multicolumn{2}{c|}{TimeXer} 
  & \multicolumn{2}{c|}{iTransformer}
  & \multicolumn{2}{c|}{RLinear}
  & \multicolumn{2}{c|}{PatchTST}
  & \multicolumn{2}{c|}{Crossformer}
  & \multicolumn{2}{c|}{TimesNet}
  & \multicolumn{2}{c|}{DLinear}
  & \multicolumn{2}{c}{Autoformer} \\
\cmidrule(l){2-3} \cmidrule(l){4-5} \cmidrule(l){6-7} \cmidrule(l){8-9} \cmidrule(l){10-11} \cmidrule(l){12-13} \cmidrule(l){14-15} \cmidrule(l){16-17} \cmidrule(l){18-19} \cmidrule(l){20-21}
  & MSE & MAE & MSE & MAE & MSE & MAE & MSE & MAE & MSE & MAE & MSE & MAE & MSE & MAE & MSE & MAE & MSE & MAE & MSE & MAE \\
\midrule
Electricity & \underline{0.168} & \underline{0.263} & \textbf{0.165} & \textbf{0.261} & 0.171 & 0.270 & 0.178 & 0.270 & 0.219 & 0.298 & 0.205 & 0.290 & 0.244 & 0.334 & 0.192 & 0.295 & 0.212 & 0.300 & 0.227 & 0.338 \\
Traffic     & \textbf{0.417} & \underline{0.278} & \underline{0.420} & \textbf{0.276} & 0.466 & 0.287 & 0.428 & 0.282 & 0.626 & 0.378 & 0.481 & 0.304 & 0.550 & 0.304 & 0.620 & 0.336 & 0.625 & 0.383 & 0.628 & 0.379 \\
Weather     & 0.247 & \underline{0.274} & \underline{0.246} & \underline{0.274} & \textbf{0.241} & \textbf{0.271} & 0.258 & 0.279 & 0.272 & 0.291 & 0.259 & 0.281 & 0.259 & 0.315 & 0.259 & 0.287 & 0.265 & 0.317 & 0.338 & 0.382 \\
Solar-Energy     & \textbf{0.232} & \textbf{0.260} & 0.235 & 0.263 & 0.470 & 0.436 & \underline{0.233} & \underline{0.262} & 0.369 & 0.356 & 0.270 & 0.307 & 0.641 & 0.639 & 0.301 & 0.319 & 0.330 & 0.401 & 0.885 & 0.711 \\
\bottomrule
\end{tabular}
}
\end{table*}

In this section, we show the forecasting performance improvement of Cross-variation Patching (DAIF-CvP) and Frequency Filtering (DAIF-F F).
We set the hyperparameters with a patch length \( P = 16 \) in DAIF-CvP and \( K = 5 \) for the top-K amplitudes in DAIF-FF. 

\textbf{DAIF vs. Original Inverted Models.}
Table \ref{tab:DAIF_all_results} compares the forecasting performance of iTransformer, inverted Mamba, and inverted MLP across three scenarios: without data augmentation (w/o aug), with Cross-variation Patching (w/ DAIF-CvP), and with Frequency Filtering (w/ DAIF-FF).

Overall, our DAIF outperforms the baseline inverted models in both short-term and long-term forecasting benchmarks.
In most cases, DAIF-CvP or DAIF-FF achieves the best or second best result.
Specifically, DAIF-CvP achieves 25 top MSEs and 24 top MAEs, while DAIF-FF achieves 14 top MSEs and 14 top MAEs. 
We note that DAIF-CvP outperforms DAIF-FF. Moreover, DAIF-CvP requires fewer input tokens for N-variate dataset when \( N > \lfloor \frac{T}{P} \rfloor \) with \(T\) lookback length.

We also analyze the impact of DAIF on different backbone models.
On one hand, we observe that DAIF enhances the iTransformer and inverted Mamba more significantly than the inverted MLP.
Overall, DAIF achieves an average MSE reduction of 3.1\% on iTransformer, 2.0\% on inverted Mamba, and 1.1\% on inverted MLP.
We believe that DAIF effectively leverages the multivariate correlation learning capabilities of inverted Seq2Seq models.
On the other hand, given that DAIF also performs well with the inverted MLP, it could be effective with other backbone models that include MLPs.
Therefore, DAIF method is considered universal and effective on inverted framework.

\textbf{DAIF vs. Competitive Models.}
To further demonstrate the contribution of DAIF, we compare its performance against several previous state-of-the-art (SOTA) models for MTS forecasting. While iTransformer has been validated as a strong SOTA baseline, we also report results for other leading methods, including TimeXer \cite{wang2024timexer}, RLinear \cite{li2023revisiting}, PatchTST \cite{nie2023time}, Crossformer \cite{zhang2023crossformer}, TimesNet \cite{wu2023timesnet}, DLinear \cite{zeng2023transformers}, and Autoformer \cite{wu2021autoformer}.

Table \ref{tab:DAIF_more_comparisons} reports the results of DAIF-CvP and DAIF-FF with iTransformer as the backbone. Overall, both methods achieve superior or comparable forecasting accuracy to previous SOTA models on the four representative datasets.
In particular, DAIF achieves the lowest MSE and MAE scores on Electricity, Traffic and Solar-Energy, while also matching or exceeding SOTA performance on Weather. 
These results highlight the generalization ability and effectiveness of DAIF in advancing MTS forecasting.

\begin{table*}[ht!]
\centering
\caption{Performance comparison of different data augmentation strategies based on iTransformer.
\textbf{Bold} highlights the best result and \underline{underline} highlights the second best result for each model.
}
\label{tab:iTransformer_DAIF_results}
\resizebox{\textwidth}{!}{%
\begin{tabular}{@{}l|cc|cc|cc|cc|cc|cc|cc@{}}
\toprule
\multirow{2}{*}{Models} & \multicolumn{2}{|c|}{w/o aug} & \multicolumn{2}{c|}{APP} & \multicolumn{2}{c|}{jitter} & \multicolumn{2}{c|}{scaling} & \multicolumn{2}{c|}{rotation} & \multicolumn{2}{c|}{\textbf{DAIF-CvP}} & \multicolumn{2}{c}{\textbf{DAIF-FF}} \\ 
\cmidrule(l){2-15}
                        & MSE & MAE & MSE & MAE & MSE & MAE & MSE & MAE & MSE & MAE & MSE & MAE & MSE & MAE \\ 
\midrule
Electricity             & 0.178        & 0.270         & 0.179        & 0.275        & 0.176        & 0.268        & 0.184        & 0.279        & 0.682        & 0.619        & \underline{0.168}        & \underline{0.263}        & \textbf{0.165} & \textbf{0.261} \\
Traffic                 & 0.428        & 0.282        & 0.423        & 0.289        & \textbf{0.414}        & 0.279        & 0.421        & 0.287        & 1.521        & 0.829        & \underline{0.417}        & \underline{0.278}        & 0.420        & \textbf{0.276} \\
Weather                 & 0.258        & 0.278        & 0.266        & 0.288        & 0.260        & 0.280        & 0.261        & 0.285        & 0.661        & 0.546        & \underline{0.247}        & \textbf{0.274}        & \textbf{0.246} & \underline{0.274} \\
\bottomrule
\end{tabular}
}
\end{table*}

\begin{table*}[ht!]
\centering
\caption{Performance comparison of inverted models with different DAIF strategies on MSE. \textbf{Bold} highlights the best result.}
\label{tab:DAIF_more_results}
\resizebox{\textwidth}{!}{%
\begin{tabular}{@{}l|ccc|ccc|ccc@{}}
\toprule
\multirow{2}{*}{MODELS} & \multicolumn{3}{|c|}{iTransformer} & \multicolumn{3}{c|}{inverted Mamba} & \multicolumn{3}{c}{inverted MLP} \\ 
\cmidrule(l){2-4} \cmidrule(l){5-7} \cmidrule(l){8-10}
                        & \textbf{DAIF-CvP} & \textbf{DAIF-FF} & \textbf{DAIF-CvP+DAIF-FF} & \textbf{DAIF-CvP} & \textbf{DAIF-FF} & \textbf{DAIF-CvP+DAIF-FF} & \textbf{DAIF-CvP} & \textbf{DAIF-FF} & \textbf{DAIF-CvP+DAIF-FF} \\ 
\midrule
Electricity             & 0.168       & \textbf{0.165}      & 0.173               & \textbf{0.220}       & 0.222      & 0.222               & \textbf{0.203}       & 0.204      & 0.203               \\
Weather                 & 0.247       & \textbf{0.246}      & 0.260               & \textbf{0.250}       & 0.263      & 0.251               & \textbf{0.271}       & 0.273      & 0.275               \\
Traffic                 & \textbf{0.417}       & 0.420      & 0.424               & 0.573       & \textbf{0.573}      & 0.574               & \textbf{0.494}       & 0.500      & 0.500               \\
\bottomrule
\end{tabular}
}
\end{table*}

\begin{figure}[ht!]	
	\subfigure[P in Cross-variation Patching.]
	{
		\begin{minipage}{4cm}
			\centering
			\includegraphics[scale=0.2]{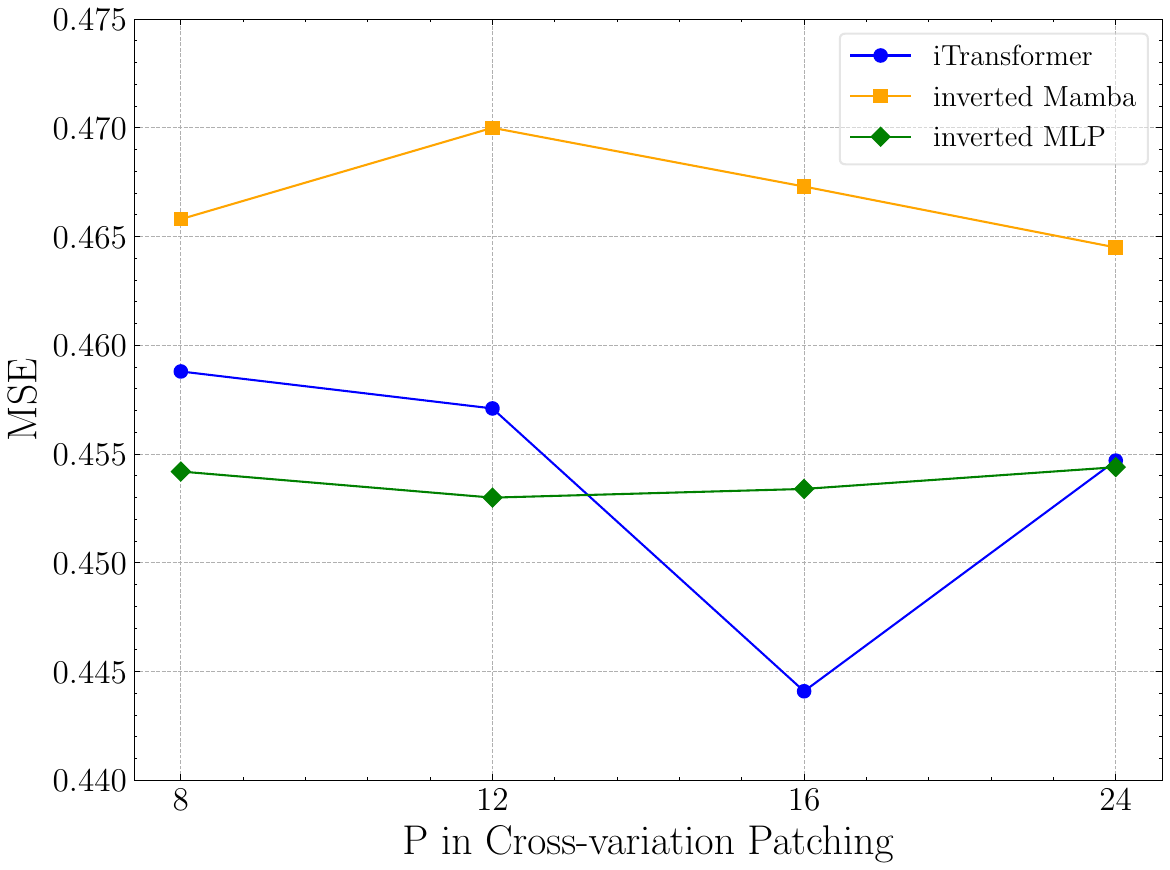}
		\end{minipage}
	}
	\subfigure[K in Frequency Filtering.]
	{
		\begin{minipage}{4cm}
			\centering
			\includegraphics[scale=0.2]{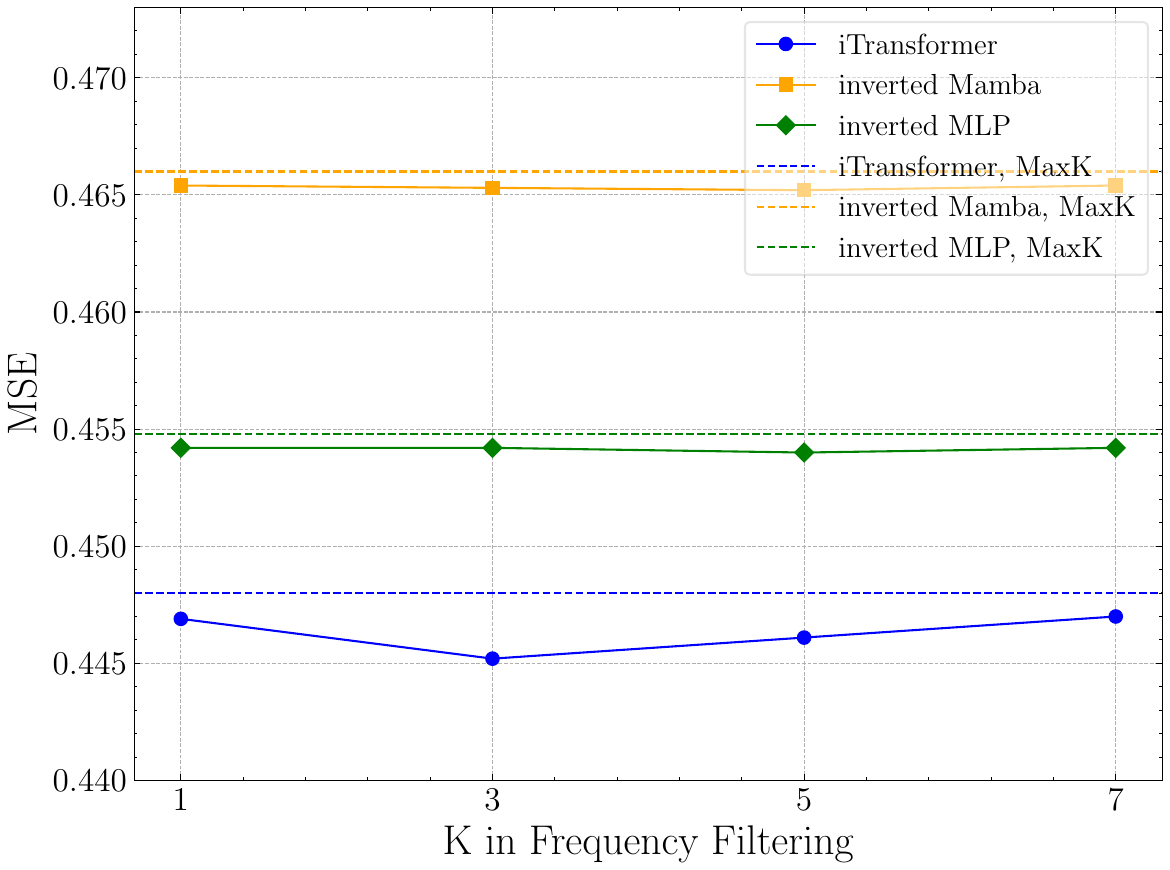}
		\end{minipage}
	}
	\caption{Sensitive analysis of hyperparameters in two DAIF strategies.}
	\label{fig:hyperparameters}
\end{figure}

\begin{figure}[ht!]
  \centering
  \subfigure[DAIF-CvP]{
    \includegraphics[width=0.8\linewidth]{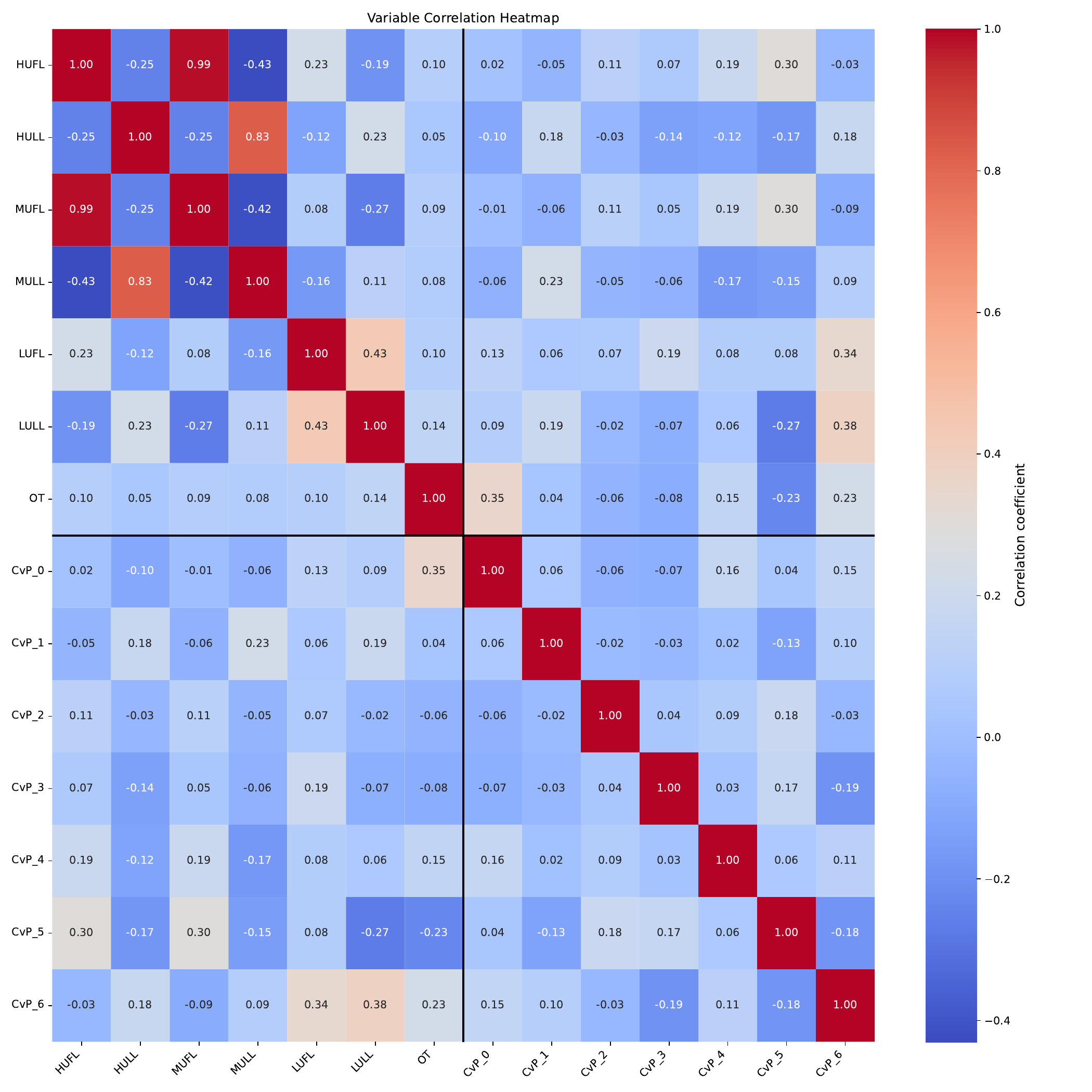}
    \label{fig:DAIF-CvP_heatmap}
  }
  \vspace{2mm}
  \subfigure[DAIF-FF]{
    \includegraphics[width=0.8\linewidth]{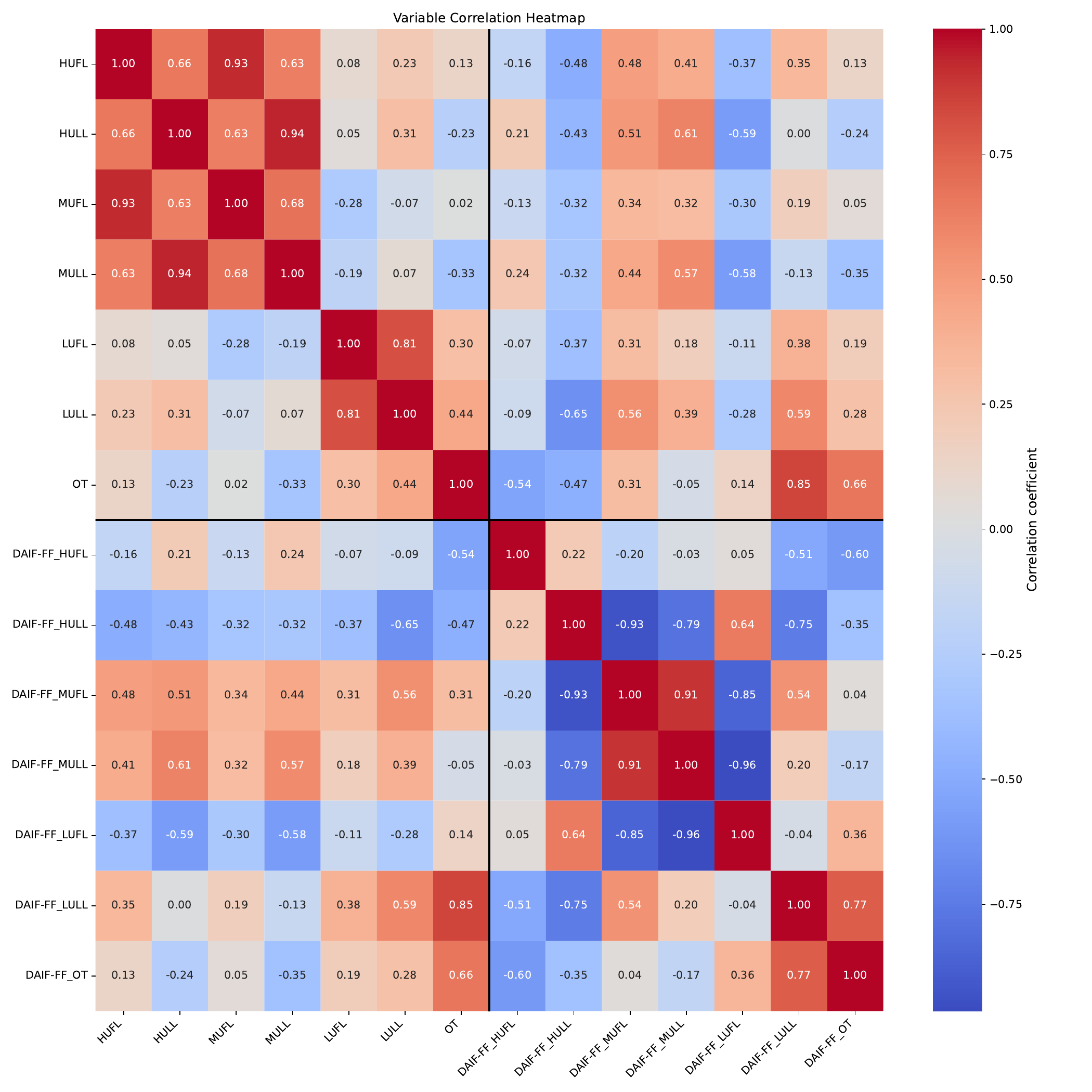}
    \label{fig:DAIF-FF_heatmap}
  }
  \caption{Variable correlation heatmaps of original and augmented variables on the ETTh1 dataset. (a) DAIF-CvP augmented variables. (b) DAIF-FF augmented variables.}
  \label{fig:daif_heatmaps}
\end{figure}

\subsection{Comparison of Different Data Augmentation Strategies}

In Table \ref{tab:iTransformer_DAIF_results}, we provide a comparison with four popular data augmentation methods: amplitude and phase perturbations (APP) \cite{gao2020robusttad}, jittering, scaling, and rotating \cite{rashid2019times}. Unlike these methods, our DAIF is the first \textbf{inverted-framework-based} data augmentation method in multivariate time series (MTS) forecasting. 
All methods were evaluated with the iTransformer as the backbone model while keeping the parameter settings consistent. The experimental results were averaged across four prediction lengths \(S \in \{96, 192, 336, 720\}\). The results in Table \ref{tab:iTransformer_DAIF_results} substantiate the advantages of our approaches (DAIF-CvP and DAIF-FF).

\subsection{Analysis of Compounded DAIF Strategies}
We experiment with the simultaneous integration of Cross-variation Patching and Frequency Filtering strategies into the inverted models (DAIF-CvP+DAIF-FF), as shown in Table \ref{tab:DAIF_more_results}.
We observe that simultaneously adding two DAIF strategies to the inverted model does not yield additional benefits.
Note that excessive data augmentation may increase the risk of losing the semantic representation of the original series.
Additionally, the integration of heterogeneous data augmentation strategies complicates interpretability.

\subsection{Analysis of Hyperparameters}
We study the impact of patch lengths \(P\) in Cross-variation Patching and frequency counts \(K\) in Frequency Filtering on the forecasting performance of three inverted models. In experiments, we employ \( P \in \{8, 12, 16, 24\} \) and \( K \in \{1, 3, 5, 7\} \) on ETTh1, with the results displayed in Fig. \ref{fig:hyperparameters}. 

Fig. \ref{fig:hyperparameters} (a) shows that inverted models are sensitive to \( P \) in Cross-variation Patching, since the patch size ultimately affects the number of DAIF input tokens, as detailed in Sec. \ref{ssec:patching}. 
Fig. \ref{fig:hyperparameters} (b) shows that inverted models have more stable performance across various values of \( K \) with the same dimensions of \( \overline{\mathbf{X}} \). 
Fig. \ref{fig:hyperparameters} (b) also presents the inverted models performance with maximum K (\(\lfloor T/2 \rfloor + 1\) in the real discrete Fourier transform) using horizontal dashed lines. 
Here, maximum K keeps all frequencies, namely using the original series as augmented data for DAIF. We see that selecting top-K frequencies performs better than keeping all frequencies.

\subsection{Variable Correlation Analysis for DAIF-FF and DAIF-CvP}
Fig. \ref{fig:daif_heatmaps} presents the variable correlation heatmaps on the ETTh1 dataset after applying DAIF-FF and DAIF-CvP. These results demonstrate that the two strategies enrich multivariate relationships in different ways. The ETTh1 dataset contains seven variables: HUFL, HULL, MUFL, MULL, LUFL, LULL, and OT. The heatmaps show the variable correlations calculated before training.

Specifically, the heatmap of DAIF-CvP shows that the generated cross-variation patch series are nearly uncorrelated, as indicated by the decorrelated lower-right block. 
This pattern increases feature diversity and orthogonality, and helps to model local temporal variations more effectively.
The off-diagonal blocks (lower-left and upper-right) indicate that the newly generated patches and the original variables are largely uncorrelated, further supporting the diversity introduced by DAIF-CvP.
On the other hand, the heatmap of DAIF-FF exhibits strong correlations in both the upper-left and lower-right corners, indicating a clear one-to-one correspondence between each variable and its enhanced version. 
Such a correlation pattern preserves dominant features and suppresses noise under weak inter-variable correlations. The off-diagonal blocks indicate that the augmented variables remain largely independent of the original series.

Overall, DAIF-FF preserves the main dependencies of the original series and reduces interference from insignificant inter-variable correlations.
DAIF-CvP enhances feature diversity and incorporates local semantic information into temporal representations, mitigating the lack of local temporal correlations in the inverted framework.
DAIF leads to a richer representation for more robust forecasting.

\section{Conclusion}
In this paper, we introduce DAIF, a novel on-the-fly data augmentation method building on the superior inverted framework for multivariate time series forecasting. DAIF improves the inverted Seq2Seq model by generating meaningful augmented data as additional input to address inherent challenges of the inverted framework. We develop Cross-variation Patching and Frequency Filtering DAIF strategies. Experiments on seven real-world datasets show that two DAIF strategies outperform the baseline inverted models. In the future, we will continue to explore more time series augmentation strategies and optimization methods on inverted framework.

\bibliographystyle{elsarticle-num}
\bibliography{reference.bib}

\end{document}